\PassOptionsToPackage{table,dvipsnames}{xcolor}
\documentclass[10pt,twocolumn,letterpaper]{article}
\usepackage[pagenumbers]{cvpr}

\usepackage[utf8]{inputenc}
\usepackage[T1]{fontenc}
\usepackage{amsfonts}
\usepackage{nicefrac}
\usepackage{microtype}
\usepackage{multirow}
\usepackage{algorithm}
\usepackage[noend]{algpseudocode}
\usepackage{pifont}
\usepackage{csquotes}
\usepackage{makecell}
\usepackage{bm}
\usepackage{array}
\usepackage{cuted}    
\usepackage{currfile}

\newcommand{\cmark}{\textcolor{green!60!black}{\ding{51}}}
\newcommand{\xmark}{\textcolor{red!75!black}{\ding{55}}}

\newcommand{\tfyes}{\textcolor{darkgreen}{\cmark}}
\newcommand{\tfno}{\textcolor{red}{\xmark}}

\definecolor{taskblue}{RGB}{214,228,249}
\definecolor{llmpink}{RGB}{246,214,214}
\definecolor{zeroyellow}{RGB}{247,238,181}
\definecolor{visiongreen}{RGB}{220,236,226}

\definecolor{ourslight}{RGB}{239,235,251}
\definecolor{oursmid}{RGB}{228,219,248}
\definecolor{oursdark}{RGB}{214,201,243}
\definecolor{twodmint}{RGB}{220,239,226}

\definecolor{darkgreen}{RGB}{0,110,40}
\newcommand{\gain}[1]{{\scriptsize\textcolor{darkgreen}{(+#1\%)}}}

\newcommand{\circnum}[1]{\raisebox{0.2pt}{\scriptsize\textcircled{\raisebox{-0.8pt}{\tiny #1}}}}

\newcommand{\ourcelllight}[1]{\cellcolor{ourslight}{#1}}
\newcommand{\ourcellmid}[1]{\cellcolor{oursmid}{#1}}
\newcommand{\ourcelldark}[1]{\cellcolor{oursdark}{#1}}
\newcommand{\ourbest}[2]{\cellcolor{oursdark}{\makecell[c]{\textbf{#1}\\[-1pt]#2}}}

\newcommand{\ours}[1]{\textcolor{teal}{\texttt{ViewAgent3D}}}
\newcommand{\judge}[1]{\textcolor{black}{\texttt{SEU}}}

\newcommand{\subtakeaway}[1]{\begin{center}\textit{--- #1}\end{center}}
\definecolor{AlgoBlue}{RGB}{30,70,230}
\algrenewcommand\algorithmiccomment[1]{\hfill{\color{AlgoBlue}// #1}}

\newcommand{\relworktablelabel}{tab:rw_positioning}

\definecolor{cvprblue}{rgb}{0.21,0.49,0.74}
\usepackage[breaklinks,colorlinks,allcolors=cvprblue]{hyperref}
\hypersetup{%
  pdftitle={ViewAgent3D: A Training-Free Multi-View Pipeline for 3D Scene Understanding},
  pdfauthor={Sarosij Bose, Arindam Dutta, Konstantinos Karydis, Amit K. Roy-Chowdhury}
}

\def\confName{CVPR}
\def\confYear{2027}
\title{\textbf{\ours{}}: A Training-Free Multi-View Pipeline for 3D~Scene~Understanding}
\author{%
  \normalsize Sarosij~Bose$^{\dagger}$\quad Arindam~Dutta$^{\dagger}$\quad
  Konstantinos~Karydis$^{\dagger}$\quad Amit~K.~Roy-Chowdhury$^{\dagger}$\\[2pt]
  \small $^{\dagger}$University of California, Riverside
}

\begin{document}
\raggedbottom
\maketitle
\begingroup
\let\viewagentarxivincludegraphics\includegraphics
\renewcommand{\includegraphics}[2][]{%
  \viewagentarxivincludegraphics[width=0.95\linewidth]{#2}}
\begin{strip}
  \centering
  \includegraphics[width=1\linewidth]{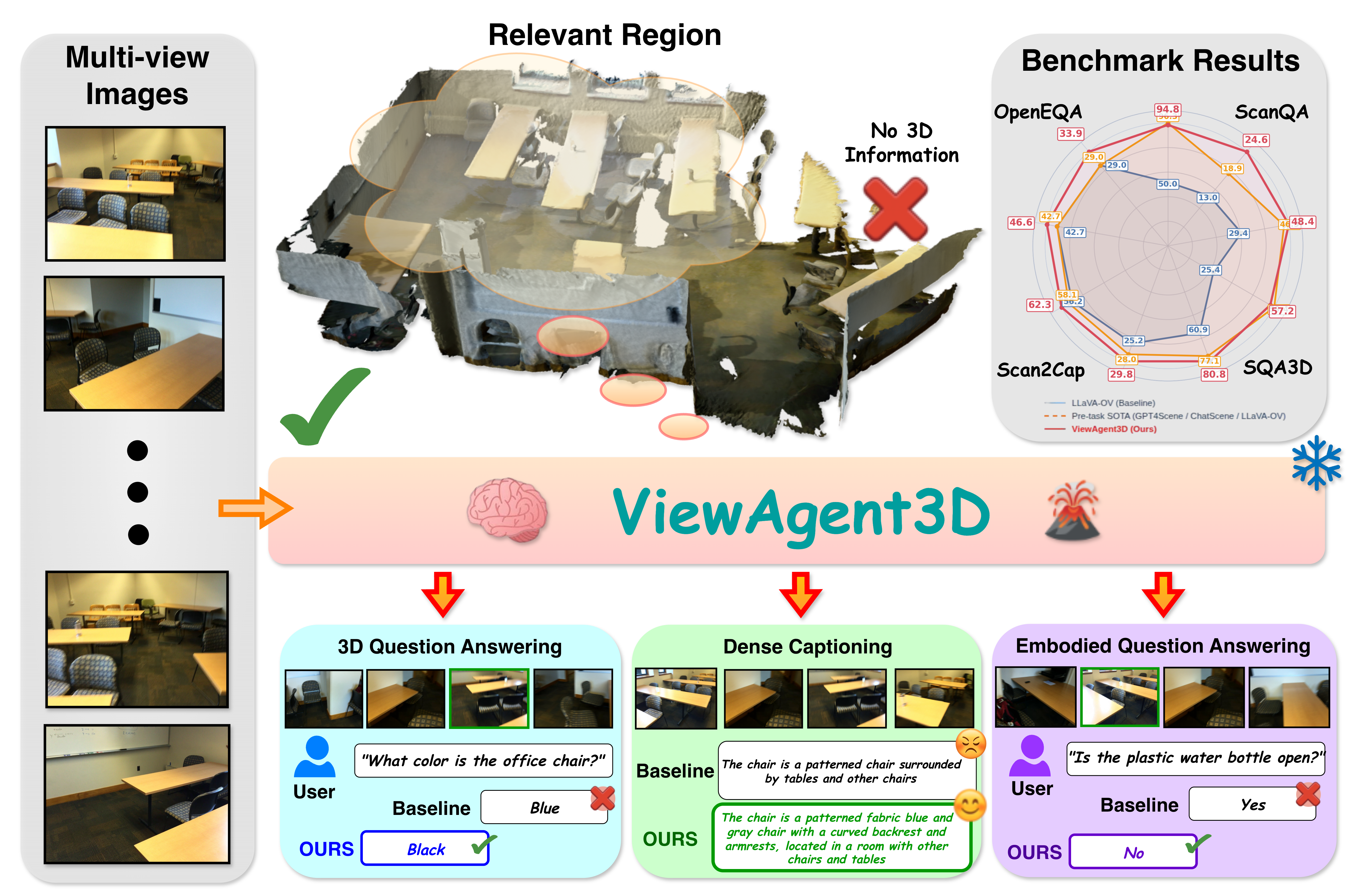}
  \captionsetup{hypcap=false}
  \captionof{figure}{%
    \textbf{Overview:} We introduce \ours{}, a \textbf{training-free, multi-view only} framework that performs image-based scene reasoning and understanding without any \textit{explicit 3D information or task-specific fine-tuning}. Experimental results show \ours{} achieves state-of-the-art performance across various downstream tasks. (Scene point cloud shown for visualization purposes only.)
  }
  \captionsetup{hypcap=true}
  \label{fig:teaser}
\end{strip}

\endgroup
\begin{abstract}
  3D scene understanding tasks are often approached by building explicit 3D scene representations or by training specialized multi-modal models over point clouds, voxels, or scene graphs. We take a complementary view: for many queries, the main bottleneck is not the absence of a stronger 3D representation but the selection of a small subset of multi-view RGB views that actually makes the query answerable. We present a training-free, query-conditioned multi-view pipeline that performs diverse tasks using only 2D images and a set of tools used purely for inference. Our method first decomposes each query into a set of concepts that identify unique attributes that assist with reasoning. This drives a coarse-to-fine evidence selection process: a coarse retrieval selects and scores all views, a diversity-aware selector reduces redundancy, and a reranking stage helps shortlist views by target visibility or answerability. The resulting evidence is optionally grounded for queries that benefit from explicit localization, and a self-verification step helps refine the final response, especially for open-ended cases. Despite using no task-specific training and no explicit 3D representations at inference time, our approach provides $\approx10\%$ gain on 3D VQA and 5\% gain on dense captioning and Embodied QA while being applicable across different scenarios. We further propose a unique judge-based evaluation system, motivating a judge who takes into account many semantically correct or partially correct answers, thereby improving holistic scene understanding.
\end{abstract}
\section{Introduction}
\label{sec:introduction}
3D reconstruction has advanced rapidly in recent years, driven in part by Neural Radiance Fields (NeRFs)~\cite{mildenhall2021nerf, kerr2023lerf} and 3D Gaussian Splatting (3DGS)~\cite{kerbl20233d, qin2024langsplat, zhou2024hugs}, enabling high-quality 3D asset generation~\cite{ruiz2023dreambooth, poole2022dreamfusion, xiang2025structured} and Simultaneous Localization And Mapping (SLAM) from dense multi-view images~\cite{matsuki2024gaussian}. 
However, in practical cases, reconstruction alone is insufficient to achieve optimal outcomes. 
An agent must not only recover scene geometry but also reason about objects, attributes, and spatial relationships to act effectively. 
This motivates the integration of Multimodal Large Language Models (MLLMs)~\cite{li2024llava, wang2024qwen2, zhang2023video} for 3D-aware spatial reasoning. 
While recent MLLMs have shown strong outcomes in 2D visual understanding~\cite{wang2024qwen2, zhang2024llava}, they still struggle to perform reliable reasoning in 3D environments~\cite{kamath2023s}. 
Prior work has primarily addressed this limitation by incorporating explicit 3D structure into the reasoning pipeline, including language-aligned 3D representations~\cite{zhu2024llava, wang2023chat, huang2023chat}, direct use of depth cues~\cite{zhu20233d, jang2025pow3r, zheng2025learning}, and joint representation learning between 3D features and language~\cite{huang2023chat, zheng2025video,thai2025splattalk}. 
Although these methods are promising, they often rely on costly 3D information~\cite{zhu20233d, yang20253d}, specialized representation supervision~\cite{leroy2024grounding, wang2024dust3r, huang2025mllms}, or reconstructed scene assets~\cite{xiang2025structured}. 
As a result, they remain brittle when handling viewpoint-dependent spatial relations, occlusions, or compositional queries that require integrating multiple aspects of evidence. 

A complementary direction leverages multi-view inputs~\cite{wang20253d}, bird's-eye-view (BEV) maps~\cite{qi2025gpt4scene}, and auxiliary scene priors~\cite{yang20253d, qi2025gpt4scene} to provide a broader spatial context. 
However, these approaches face persistent trade-offs because they must preserve global scene understanding while retaining the fine-grained local evidence needed to accurately answer a particular question. 
In contrast to these existing approaches, we ask an orthogonal question: \textit{Can an off-the-shelf MLLM reason about a 3D scene using only multi-view 2D observations without necessitating 3D reconstruction?} 
Instead of assuming that reliable 3D reasoning requires an intermediate 3D representation, we investigate whether 3D visual reasoning can be recovered from 2D evidence alone. 
Our central premise is that spatial queries may be answerable when the system can retrieve the appropriate viewpoints, isolate the relevant visual evidence, and integrate this evidence through robust multimodal reasoning. 
As shown in {\hyperref[fig:teaser]{Figure~\ref{fig:teaser}}}, even a seemingly simple question such as \emph{``What color is the office chair?''} requires identifying the relevant entities and spatial relations, identifying views in which they are jointly visible, and reasoning over the collected information. 

To address these challenges, \ours{} is a \textit{training-free, test-time reasoning pipeline} for 3D scene understanding over multiview image observations of a scene. 
Rather than necessitating explicit 3D reconstruction or point clouds per scene, \ours{} tackles \emph{query-conditioned information selection} as the core problem. 
Given a natural-language query and a set of multi-view images, \ours{} first parses it using a \textbf{Concept Decomposition} step to extract relevant semantic components, including target objects, attributes, and relations, as shown in Figure~\ref{fig:pipeline}. 
\ours{} then performs a \textbf{View Selection} process to perform coarse retrieval, producing a compact set of candidate views conditioned on the input. 
This step is crucial because, for most questions, the majority of frames are irrelevant, and many relevant frames may be redundant. 
Over this set, to remove redundant views and increase diversity, we mix the retrieved views with diverse ones to form a well-informed set of views for the query. 
Next, a \textbf{VLM Ranking step} is used to perform a visibility-aware ranking to identify which views actually contain the evidence needed to answer the question. 
To aid with this step, a \textbf{Conditional Grounding} tool is invoked selectively only when explicit localization is likely to disambiguate the relevant concepts. 
This process ends with a self-verification step that generates the final answer when the curated information is extracted from the selected views. 
We further introduce \texttt{SEU}, a holistic judge with metrics for open-ended cases where a prediction differs lexically from the ground truth but preserves the \mbox{same meaning}. 

\begin{table*}[t]
\centering
\footnotesize
\caption{\textbf{Comparison of \ours{} relative to representative prior methods.}
\ours{} is the only method here that is training-free, 3D-agnostic at inference time, explicitly reasoning-centric, and free from additional scene geometry such as BEV maps or~geometric~markers.}
\label{\relworktablelabel}

\begin{tabular}{@{}lccccc@{}}
\toprule
\textbf{Method} & \textbf{Training-Free} & \textbf{3D-Agnostic} & \textbf{RGB Inference} & \textbf{BEV / Marker Free} & \textbf{Reasoning / Tool Use} \\
\midrule
ChatScene~\cite{huang2023chat}       & \xmark & \xmark & \xmark & -- & \xmark \\
LL3DA~\cite{chen2024ll3da}           & \xmark & \xmark & \xmark & -- & \xmark \\
CdViews~\cite{wang20253d}            & \xmark & \cmark & \cmark & \cmark & \xmark \\
SplatTalk~\cite{thai2025splattalk}   & \xmark & \xmark & \cmark & \cmark & \xmark \\
GPT4Scene~\cite{qi2025gpt4scene}     & \cmark & \cmark & \cmark & \xmark & \xmark \\
\midrule
\rowcolor{oursmid}
\textbf{\ours{}}                     & \textbf{\cmark} & \textbf{\cmark} & \textbf{\cmark} & \textbf{\cmark} & \textbf{\cmark} \\
\bottomrule
\end{tabular}
\end{table*}

Extensive experiments across 3D VQA, dense captioning, and Embodied QA show that \ours{} achieves strong performance despite being training-free and RGB-only. 
By extracting compact, query-relevant information, our method substantially outperforms previous video-based approaches and surpasses several methods that incorporate explicit 3D information. 
On 3D QA benchmarks, \ours{} improves over existing approaches by \textbf{9}\% and \textbf{5}\% relatively on average, outperforming 3D-pretrained models such as ChatScene~\cite{huang2023chat} and 2D-trained models such as SplatTalk~\cite{thai2025splattalk}, while improving the underlying backbone VLM by more than \textbf{25}\%. 
These results suggest that selecting accurate 2D information from views can often compensate for the absence of explicit 3D representations, offering a simple yet powerful design principle for 3D scene understanding. Table~\ref{tab:rw_positioning} provides a comparison w.r.t to \mbox{representative prior works}.

In summary, our main contributions are as follows:
\begin{itemize}
    \item We introduce \ours{}, a \emph{training-free, 2D-only} pipeline for 3D scene understanding tasks that reasons directly over multi-view RGB observations without the need to construct an explicit 3D scene representation.

    \item We propose a query-driven reasoning and tool utilization framework that combines concept extraction, retrieval of views, and relevance-diversity balanced viewpoint selection, visibility-aware ranking, selective grounding, and self-verified response generation.

    \item We introduce a holistic evaluation setup using a Judge-based mechanism that improves evaluation pertaining to open-ended cases by evaluating responses more holistically rather than deterministic metric evaluation, and improves recovery by around ~\textbf{15}\%.

    \item We demonstrate strong performance on diverse tasks, including 3D QA benchmarks, dense captioning, and Embodied QA, providing gains by over $\sim$\textbf{10}\% on average, all while without any task-specific fine-tuning or usage of any 3D information.
\end{itemize}

\section{ViewAgent3D}
\label{sec:method}
\begin{figure*}[t]
    \centering
    \includegraphics[width=1\linewidth]{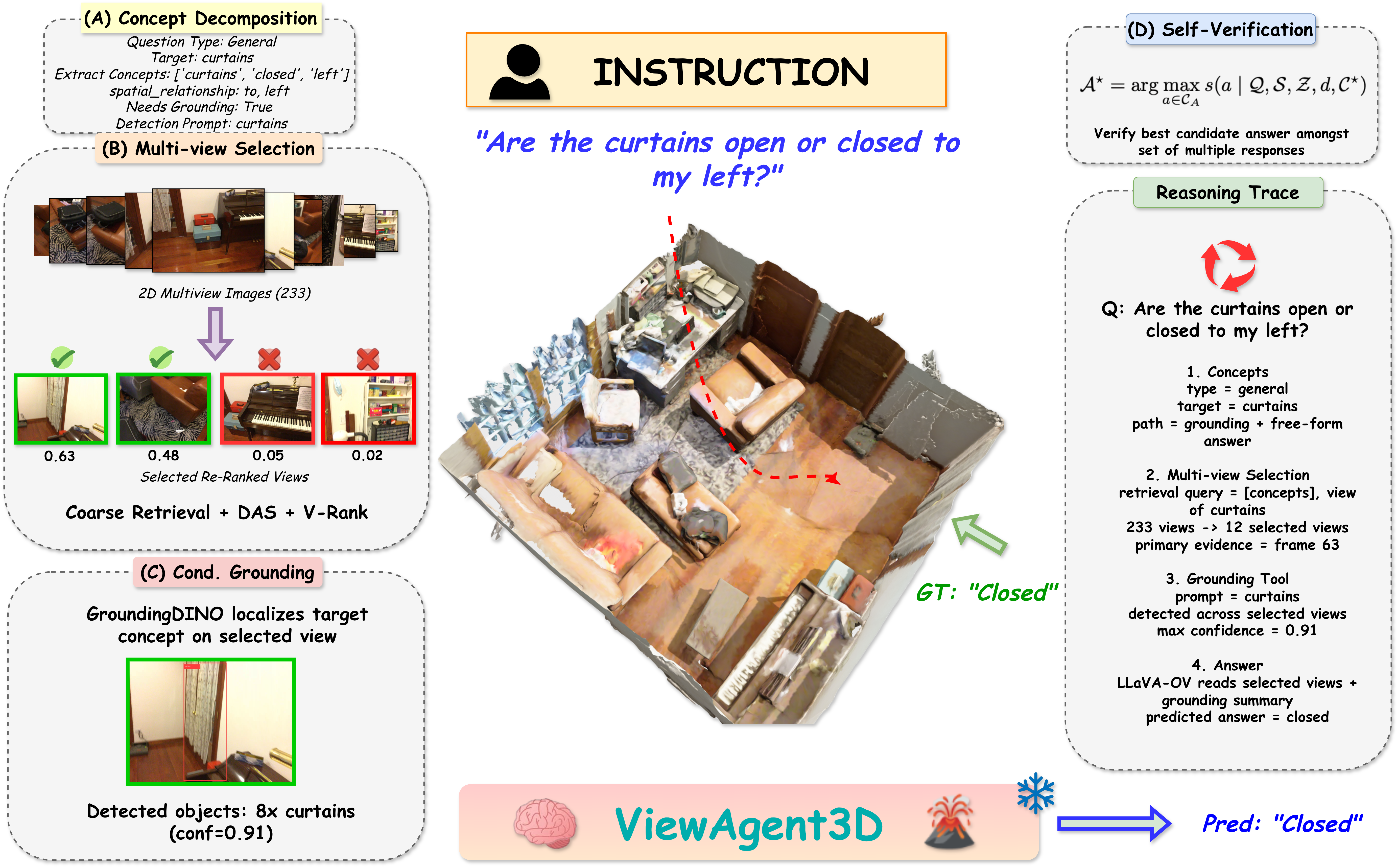}
    \caption{\textbf{Workflow Overview.} \ours{} takes in as input multi-view images $\mathcal{V}$ of the scene and the query $\mathcal{Q}$. The concept decomposition step \textbf{(A)} extracts the relevant target-related concepts from the query. Next, we narrow down the pool of viewpoints by retrieving view, increase diversity to reduce redundancy and rank them by answerability \textbf{(B)} aided by a conditional grounding tool \textbf{(C)}. After the self-verification step \textbf{(D)}, the model generates the response $\mathcal{A}$*.}
    \label{fig:pipeline}
\end{figure*}

An illustration of \ours{} is shown in {\hyperref[fig:pipeline]{Figure~\ref{fig:pipeline}}}. 
Our method treats language-driven 3D scene understanding as a \emph{test-time reasoning problem} over RGB views: instead of utilizing an explicit 3D scene representation, it reasons about which informative subset of observations makes the query answerable. 
To this end, Sec.~\ref{sec:routing} abstracts the query into a compact semantic decomposition that exposes the entities, relations, and cues the system should reason over; Sec.~\ref{sec:view_selection} uses this structure to retrieve candidate views, remove redundancy, and allow the VLM to observe and reason over views that are actually answerable; Sec.~\ref{sec:grounding} invokes 2D grounding only when explicit localization is likely to sharpen the evidence; and Sec.~\ref{sec:answering} produces the final response from the selected evidence. 
This enables the VLM to not only answer but also reason and decide which evidence can be useful before generating a response.

Formally, we represent the 3D scene as a set of multi-view RGB images \(\mathcal{V}=\{\mathcal{I}_i\}_{i=1}^{N}\), where \(\mathcal{I}_i \in \mathbb{R}^{H \times W \times 3}\) denotes the \(i\)-th RGB image. Given a query \(\mathcal{Q}\) and optional context \(\mathcal{S}\), the goal is to predict a textual output from the VLM \(\mathcal{F}(\cdot)\),
\begin{equation}
    \mathcal{A}=\mathcal{F}(\mathcal{V},\mathcal{Q},\mathcal{S}).
\end{equation}
The remainder of the method specifies how \ours{} reasons over $\mathcal{V}$ through tool-based evidence selection and uses the resulting information to generate the final response.
\subsection{Semantic Concept Decomposition}
\label{sec:routing}
\subtakeaway{Extracts a compact set of concepts that guide retrieval, grounding, and answer generation.}
Inspired by semantic parsing and symbolic visual reasoning techniques~\cite{johnson2017inferring, yi2018neural}, we first extract the relevant set of concepts from the query $\mathcal{Q}$ into a compact set $\mathcal{Z}$,
\begin{equation}
    \mathcal{Z}=\Phi(\mathcal{Q})=\bigl(c,t,r,\rho,\alpha,\gamma,\kappa\bigr),
    \label{eq:concept_state}
\end{equation}
where $c$ denotes the question type, $t$ the target concept, $r$ an optional reference concept, $\rho$ an optional relation, $\alpha$ an attribute cue, $\gamma \in \{0,1\}$ a grounding flag, and $\kappa \in [0,1]$ a parse confidence score. 
The role of $\mathcal{Z}$ is to extract only the semantic factors that later stages need to reason over: what the query is about, whether disambiguation requires a reference object, and whether explicit grounding is likely to help. 
This structured set is particularly useful for compositional queries, where retrieval over raw text can drift toward incidental nouns or views with similar salient features. 
For example, in \emph{“What shape is the refrigerator on the left of a kitchen counter?”}, $\Phi$ isolates the target (\emph{refrigerator}), reference (\emph{kitchen counter}), and relation (\emph{left of}), yielding a cleaner interface for evidence acquisition.

\subsection{Multi-view Evidence Selection}
\label{sec:view_selection}
\subtakeaway{The central reasoning step is selecting views that are answerable, not merely relevant.}
We formulate the retrieval query $\mathcal{G}$ with a view budget $\mathcal{K}$ augmented with the obtained semantic decomposition~set~$\mathcal{Z}$~\cite{prasad2023rephrase},
\begin{equation}
    \mathcal{G}=G(\mathcal{Q},\mathcal{S},\mathcal{Z}), \qquad
    \mathcal{B}(\mathcal{Z})=(K_{\mathcal{Z}},M_{\mathcal{Z}}),
\end{equation}
where $\mathcal{B}(\mathcal{Z})$ denotes the evidence budget, with $M_{\mathcal{Z}}$ as the shortlist size and $K_{\mathcal{Z}}$ as the final number of views passed to the answering model $\mathcal{F(\cdot)}$. 
Thus, the query $\mathcal{G}$ encodes a structured relationship between the target, reference, and relation given in $\mathcal{Z}$ rather than directly processing open-ended queries with an adaptive budget for selecting views (i.e. smaller for attribute/object-centric queries and larger for location or referential, and general queries).

\textbf{Coarse candidate view retrieval.} We begin with a fast, high-recall retrieval stage that reduces $N$ multi-view images to a smaller candidate pool. 
Specifically, we encode the structured query ($\mathcal{G} \rightarrow \mathbf{e}_{\mathcal{Q}}$) and each image ($\mathcal{I}_i \rightarrow \mathbf{e}_i$) into a shared embedding space,
\begin{equation}
    \mathbf{e}_{\mathcal{Q}} = f_t(\mathcal{G}) \in \mathbb{R}^{d}, 
    \qquad
    \mathbf{e}_i = f_v(\mathcal{I}_i) \in \mathbb{R}^{d},
\end{equation}
using a SigLIP2 retriever~\cite{tschannen2025siglip}, and score every view by,
\begin{equation}
    u_i=\langle \mathbf{e}_{\mathcal{Q}}, \mathbf{e}_i\rangle,\qquad i=1,\ldots,N.
    \label{eq:siglip_score}
\end{equation}
Here, $\langle\cdot,\cdot\rangle$ denotes the inner product with $\ell_2$-normalized embeddings, and $u_i$ corresponds to the cosine similarity score between the query and view representations. 
We then retain the top-$M_{\mathcal{Z}}$ scored views as the initial candidate pool. 
This formulation intentionally selects relevant frames within the broader scene context while removing undesirable frames; however, it is still insufficient to determine whether the selected views contain sufficient visual evidence to answer the query. 
Because the image embeddings $\{\mathbf{e}_i\}_{i=1}^{N}$ depend only on the scene, they can be computed once and reused across multiple queries.

\textbf{\underline{D}iversity-\underline{A}ware View \underline{S}election (DAS).}
The retrieval of views from the previous step is often coarse and leads to clustering around the same object or viewpoint. 
This leads to redundancy in the selected views, which hinders effective reasoning: overlapping frames increase the cost without contributing any new information. 
Therefore, DAS reduces the retrieved views into a refined set of views that is both relevant to the query and reduces overlap across viewpoints. 
To measure redundancy between two candidate views \(i\) and \(j\), we combine a unit-scale viewpoint distance $d_{ij}$ with the appearance similarity represented by the image embeddings,
\begin{equation}
    \mathrm{sim}(i,j)=
    \eta\,\exp\!\bigl(-d_{ij}/\sigma\bigr)
    +(1-\eta)\,\max\!\bigl(0,\,\langle \mathbf{e}_i,\mathbf{e}_j\rangle\bigr),
    \label{eq:sim}
\end{equation}
where \(\eta\in[0,1]\) balances the viewpoint-level and appearance-level redundancy, and \(\sigma\) is a decay constant. 
Intuitively, \(\mathrm{sim}(i,j)\) is large when two views contain near-duplicate target concepts or appearances. 
Using this redundancy score together with retrieval relevance, we greedily build the set \(\mathcal{C}\) with Maximal Marginal Relevance (MMR)~\cite{carbonell1998use}. 
Let \(\hat{u}_i=\frac{u_i-\min_{j}(u_j)}{\max_{j}(u_j)-\min_{j}(u_j)}\) denote the normalized retrieval score for the candidate set. 
At each step, DAS selects the next view \(i^\star\) by,
\begin{equation}
    i^\star=\operatorname{\mathbf{arg\,max}}\nolimits_{i\notin\mathcal{C}}
    \left[
        \lambda \hat{u}_i-(1-\lambda)\max_{j\in\mathcal{C}}\bigl(\mathrm{sim}(i,j)\bigr)
    \right],
    \label{eq:mmr}
\end{equation}
where \(\lambda\in[0,1]\) controls the trade-off between retrieval relevance and diversity. 
The first term preserves semantic compatibility with the query, whereas the second suppresses views that duplicate information already present in \(\mathcal{C}\). 
Balancing between the full set of views and retrieved views allows DAS to account for moderately relevant views whose information complements the retrieved ones, especially where naive top-\(M\) retrieval tends to cluster.

\textbf{\underline{V}isibility-aware VLM \underline{R}anking (V-Rank).}
After DAS, the candidate set is diverse and semantically relevant; however, some views may still be visually insufficient: the target can be occluded, too small, or not jointly visible with the reference target. 
We pass this compact set to the VLM $\mathcal{F(\cdot)}$ and use it to assess which views provide the strongest evidence for answering a query. 
Given \((\mathcal{Q},\mathcal{S})\), the semantic decomposition \(\mathcal{Z}\), and the candidate views \(\mathcal{C}\), the VLM ranks them by \emph{answerability} rather than by semantic similarity alone. 
We present the views in \(\mathcal{C}\) as a numbered set and prompt the VLM to return an ordered list \(\mathcal{R}=(i_1,\dots,i_L)\) of the most-answerable candidates. We convert this ordering into a score,
\begin{equation}
    r_i=
    \begin{cases}
        \frac{L-k+1}{L}, & i=i_k,\ k\in\{1,\dots,L\},\\
        0, & \text{otherwise},
    \end{cases}
    \label{eq:rerank_score}
\end{equation}
Therefore, higher-ranked views receive larger values. 
We then fuse this answerability score with the normalized retrieval prior \(\hat{u}_i\),
\begin{equation}
    \tilde{u}_i=(1-\lambda_r)\hat{u}_i+\lambda_r r_i,
    \qquad \lambda_r\in[0,1],
    \label{eq:fused_score}
\end{equation}
where \(\lambda_r\) controls the trade-off between semantic relevance and answerability. 
This step on \(\tilde{u}_i\) yields the selected set of views \(\mathcal{C}^{\star}\) of size \(K_{\mathcal{Z}}\). 
Thus, \textbf{V-Rank} is the main reasoning step in \ours{}. 
Instead of passing all retrieved views to the VLM, we first use it to identify the few observations that actually support the query. 
For relation-heavy questions, for example, only a subset of views may jointly expose the target, the reference, and their spatial relation clearly enough to justify an answer.

\subsection{Conditional 2D Grounding}
\label{sec:grounding}

Grounding is useful when explicit localization is likely to sharpen the evidence in views, primarily for counting, existence, and broad object-centric queries. 
Therefore, it acts as a tool for \textbf{V-Rank}, as it decides \emph{which views} are trustworthy, while grounding identifies \emph{where within those views} the target concept lies. 
When the grounding flag in \(\mathcal{Z}\) is active, we form an open-vocabulary detector prompt from the extracted target and reference concepts and apply GroundingDINO~\cite{liu2024grounding} to each selected view in \(\mathcal{C}^{\star}\),
\begin{equation}
    \mathcal{B}_i=\{(b_{ij},\ell_{ij},s_{ij})\}_j,
\end{equation}
where \(b_{ij}\) is a bounding box, \(\ell_{ij}\) is its label, and \(s_{ij}\) is its confidence. 
The answer model then receives the selected views, together with optional box overlays and a compact detection summary $\mathcal{B}$. 

\subsection{Self Verification Based Answer Generation}
\label{sec:answering}
\subtakeaway{The model first proposes an answer, then self-verifies it.}

Even after ranking, multiple plausible responses may still be possible for the same piece of information. 
Recent self-refinement methods~\cite{madaan2023self, gou2023critic} have shown that adding bounded self-verification steps under fixed evidence often helps improve the final response. 
Let \(\mathcal{A}=\mathcal{F}\!\left(P(\mathcal{Q},\mathcal{S},\mathcal{Z},d),\mathcal{C}^{\star}\right)\) denote the initial response. 
To resolve ambiguity with the same information, we allot an iterative budget \((k \le \tau)\), where \(\tau\) controls the maximum number of alternative candidates considered during the verification. 
We then ask the VLM to compare and verify this bounded candidate set under fixed information,
\begin{equation}
\mathcal{C}_{A}
=
\{\mathcal{A}\}
\cup
\left\{
\widetilde{\mathcal{A}}_{k}
:\;
\widetilde{\mathcal{A}}_{k}\neq \mathcal{A},\ k\in\{1,\dots,\tau\}
\right\},
\label{eq:candidate_set}
\end{equation}
\begin{equation}
\mathcal{A}^{\star}
=
\arg\max_{a\in \mathcal{C}_{A}}
s\!\left(a \mid \mathcal{Q},\mathcal{S},\mathcal{Z},\mathcal{B},\mathcal{C}^{\star}\right),
\label{eq:answer_select}
\end{equation}
where \(s(\cdot)\) is a scalar verification score produced by the same VLM from a single verification prompt over the bounded candidate set under the fixed selected evidence \(\mathcal{E}^{\star}\), and higher values indicate stronger support from the available observations. 
For example, the same evidence may support both \emph{window} and the more specific \emph{glass}. 
The model compares a small set of evidence-compatible candidates and commits to the one that is better supported by the selected observations. 
This enables adherence to the query more accurately and generates the final response $\mathcal{A}^{\star}$.


\begin{table*}[t]
\centering
\small
\caption{\textbf{Quantitative comparison with prior methods for 3D QA:} We evaluate on ScanQA~\cite{azuma2022scanqa}(val) and SQA3D~\cite{ma2022sqa3d}(test). \textbf{Modality} indicates whether the method is primarily point-cloud based (\textbf{PC}) or image-based (\textbf{I}). \textbf{Training-Free (TF)} indicates whether the method operates without task-specific training. We report \textbf{CIDEr}$\uparrow$~\cite{vedantam2015cider}, \textbf{METEOR}$\uparrow$~\cite{banerjee2005meteor}, \textbf{ROUGE-L}$\uparrow$~\cite{lin2004rouge}, and \textbf{Exact Match@1}$\uparrow$~\cite{antol2015vqa}. (\(\dagger\): different backbone/HD resolution\textasciitilde1.7x more info per view and BEV map.)}
\label{tab:main_vqa}

\begin{tabular}{@{}l@{\hspace{-4pt}}c@{\hspace{2.5pt}}c@{\hspace{2.2pt}}cccc c@{}}
\toprule
\multirow{2}{*}{\textbf{Method}} & \multirow{2}{*}{\textbf{Modality}} & \multirow{2}{*}{\textbf{TF}} & \multicolumn{4}{c}{\textbf{ScanQA (val)}} & \multicolumn{1}{c}{\textbf{SQA3D (test)}} \\
\cmidrule(lr){4-7}\cmidrule(l){8-8}
& & & \textbf{CIDEr}$\uparrow$ & \textbf{METEOR}$\uparrow$ & \textbf{ROUGE-L}$\uparrow$ & \textbf{EM@1}$\uparrow$ & \textbf{EM@1}$\uparrow$ \\
\midrule

\rowcolor{taskblue}
\multicolumn{8}{@{}l@{}}{\textbf{\textit{Task-Specific}}} \\
ScanQA~\cite{azuma2022scanqa}                 & PC & \tfno & 64.9 & 13.1 & 33.3 & 21.1 & --   \\
SQA3D~\cite{ma2022sqa3d}                      & PC & \tfno & --   & --   & --   & --   & 46.6 \\
3D-VLP~\cite{jin2023context}                  & PC & \tfno & --   & 13.5 & 34.5 & 21.7 & --   \\
3D-VisTA~\cite{zhu20233d}                     & PC & \tfno & 69.6 & 13.9 & 35.7 & 22.4 & 48.5 \\

\midrule
\rowcolor{llmpink}
\multicolumn{8}{@{}l@{}}{\textbf{\textit{3D LLMs}}} \\
3D-LLM~\cite{hong20233d}                      & PC+I & \tfno & 69.4  & 14.5 & 35.7 & 20.5 & --   \\
LL3DA~\cite{chen2024ll3da}                    & PC   & \tfno & 76.8  & 15.9 & 37.3 & --   & --   \\
Chat-3D v2~\cite{wang2023chat}                & PC   & \tfno & 53.2  & 11.9 & 28.5 & --   & --   \\
LEO~\cite{huang2023embodied}                  & PC+I & \tfno & 80.0  & 16.2 & 39.3 & 21.5 & 50.0 \\
Scene-LLM~\cite{fu2024scene}                  & PC+I & \tfno & 80.0  & 15.8 & 35.9 & 27.2 & 53.6 \\
ChatScene~\cite{huang2023chat}                & PC+I & \tfno & 87.7  & 18.0 & 41.6 & 21.6 & 54.6 \\
Grounded 3D-LLM~\cite{chen2024grounded}       & PC+I & \tfno & 72.7  & --   & --   & --   & --   \\

\midrule
\rowcolor{twodmint}
\multicolumn{8}{@{}l@{}}{\textbf{\textit{2D VLMs}}} \\
CdViews*~\cite{wang20253d}                    & I  & \tfno & 87.2  & 17.0 & 40.9 & 24.7 & -- \\
SplatTalk~\cite{thai2025splattalk}            & I  & \tfno & 77.5  & 15.6 & 38.5 & 22.4 & 47.6 \\
\rowcolor{gray!12}
\textcolor{gray!75!black}{GPT4Scene-HDM$^\dagger$~\cite{qi2025gpt4scene}} &
\textcolor{gray!75!black}{I} &
\textcolor{gray!75!black}{\tfno} &
\textcolor{gray!75!black}{96.3} &
\textcolor{gray!75!black}{18.9} &
\textcolor{gray!75!black}{46.5} &
\textcolor{gray!75!black}{--} &
\textcolor{gray!75!black}{59.4} \\

\midrule
\rowcolor{zeroyellow}
\multicolumn{8}{@{}l@{}}{\textbf{\textit{Zero-shot VideoLMs}}} \\
VideoChat2~\cite{li2024mvbench}               & I & \tfyes & 49.2 & 9.5 & 28.2 & 19.2 & 37.3 \\
Qwen2-VL-7B~\cite{wang2024qwen2}              & I & \tfyes & 53.9 & 11.4 & 29.3 & --   & 40.7 \\
LLaVA-NeXT-Video~\cite{li2024llava}           & I & \tfyes & 46.2 & 9.1 & 27.8 & 18.7 & 34.2 \\
LLaVA-OV~\cite{li2024llava} \textbf{(Baseline)} & I & \tfyes & 50.0 & 13.0 & 29.4 & 15.6  & 25.4 \\

\midrule
\ourcelllight{\ours{} w/o \textbf{DAS}} &
\ourcelllight{I} &
\ourcelllight{\tfyes} &
\ourcelllight{84.3} &
\ourcelllight{21.9} &
\ourcelllight{44.6} &
\ourcelllight{25.8} &
\ourcelllight{50.9} \\

\ourcellmid{\ours{} w/o \textbf{V-Rank}} &
\ourcellmid{I} &
\ourcellmid{\tfyes} &
\ourcellmid{88.6} &
\ourcellmid{22.8} &
\ourcellmid{45.9} &
\ourcellmid{26.9} &
\ourcellmid{52.8} \\

\ourcelldark{\textbf{\ours{} (Ours)}} &
\ourcelldark{\textbf{I}} &
\ourcelldark{\tfyes} &
\ourbest{94.8}{\gain{7.1}} &
\ourbest{24.6}{\gain{6.6}} &
\ourbest{48.4}{\gain{6.8}} &
\ourbest{29.3}{\gain{2.1}} &
\ourbest{57.2}{\gain{2.6}} \\
\bottomrule
\end{tabular}
\end{table*}
\section{Experiments}
\label{sec:experiments}
\label{sec:exp_setup}
\textbf{Datasets.} We evaluate \ours{} on standard 3D QA benchmarks ScanQA~\cite{azuma2022scanqa} and SQA3D~\cite{ma2022sqa3d} (Section~\ref{sec:exp_vqa}) and evaluate its cross-task transfer capabilities on Dense Captioning (Scan2Cap benchmark~\cite{chen2021scan2cap}) and Embodied Question Answering (EQA) on OpenEQA~\cite{majumdar2024openeqa} in Section~\ref{sec:exp_transfer}. Unless otherwise noted, the same \ours{} pipeline is used across all tasks in a Training-Free (TF) manner. More datasets, implementations, and default settings are presented in Supplementary Section~B.
\textbf{Evaluation Metrics.} For VQA and dense captioning, we report the Exact Match (EM)~\cite{antol2015vqa} together with METEOR~\cite{banerjee2005meteor}, ROUGE-L~\cite{lin2004rouge}, and CIDEr~\cite{vedantam2015cider} to capture lexical overlap and generation quality. For Embodied QA, we follow prior work and report LLM-Match~\cite{liu2024mmbench} and Token-F1~\cite{rajpurkar2016squad}.\\ 
\textbf{Judge-based evaluation.} Exact-match and n-gram metrics~\cite{antol2015vqa,papineni2002bleu} often fail to capture the holistic meaning needed for open-ended 3D tasks, where valid responses may differ from the ground truth only by harmless variations in specificity or phrasing (e.g., \emph{chair} vs. \emph{armchair} in {\hyperref[fig:main_result]{Figure~\ref{fig:main_result}}}, more examples in Supplementary Section~D). To capture these nuances, we introduce \judge{}, a semantic judge instantiated with GPT-4o~\cite{achiam2023gpt}. For prediction $\hat{a}_i$ and reference set $\mathcal{A}_i$, the judge assigns $y_i \in \{\texttt{correct},\texttt{partial},\texttt{error}\}$, where \texttt{correct} denotes semantic equivalence to a reference, \texttt{partial} denotes meaningful but incomplete or underspecified content, and \texttt{error} denotes an incorrect answer. 
We define Judge Correctness (JC) and Judge Partial Correct (JCP),
\begin{equation}
    \begin{aligned}
    \mathrm{JC}&=\frac{1}{N}\sum_i [\,y_i=\texttt{correct}\,],\\
    \mathrm{JCP}&=\frac{1}{N}\sum_i [\,y_i\neq\texttt{error}\,].
    \end{aligned}
\end{equation}
To quantify the semantic gap between strict and relaxed correctness, we define Semantic Recovery $\mathrm{SR} = \mathrm{JCP} - \mathrm{JC}$ and summarize the judge labels via a composite score,
\begin{equation}
    \mathrm{\judge{}_{\alpha}}
    =
    \alpha\,\mathrm{JC}
    +(1-\alpha)\,\mathrm{JCP},
    \qquad \alpha\in[0,1].
\end{equation}
We set $\alpha=0.5$, balancing semantic correctness and acceptability. Examples, ablation, full prompt details, rubric design, and human-preference validation are provided in Supplementary Sections~D and~E.
\newline\textbf{Baselines.} We compare \ours{} against multiple families of methods: \textbf{(1) Task-specific 3D methods.} trained specifically for benchmark-specific tasks, such as MCAN and 3D-VisTA~\cite{yu2019deep, zhu20233d}. \textbf{(2) 3D LLMs.} which incorporate explicit 3D scene representations or 3D-pretrained backbones; such as ChatScene~\cite{huang2023chat}, LEO~\cite{huang2023embodied}, LL3DA~\cite{chen2024ll3da}, etc. \textbf{(3) 2D LLMs.} utilizing multi-view/BEV information but not training free~\cite{wang20253d, thai2025splattalk, qi2025gpt4scene} and our baseline LLaVA-OneVision~\cite{li2024llava}. \textbf{(4) Zero-shot VLMs.} consisting of video-based baselines~\cite{zhang2024llava, li2024mvbench, wang2024qwen2}. 
For transfer experiments, we further compare against task-specific baselines on dense captioning and embodied QA, while keeping the \ours{} pipeline fixed. Additional results and ablations are provided in Supplementary Sections~C and~G.

\subsection{\ours{}: 3D Question Answering}
\label{sec:exp_vqa}
We evaluate \ours{} on 3D QA benchmarks ScanQA~\cite{azuma2022scanqa} and SQA3D~\cite{ma2022sqa3d} which are shown in {\hyperref[tab:main_vqa]{Table~\ref{tab:main_vqa}}}. Following prior work~\cite{wang20253d, thai2025splattalk} we group the methods into three categories: \textit{“Task-Specific Models"}, specifically fine-tuned for benchmarks without generalization to other tasks, \textit{“3D MLLMs”} that use point clouds or intermediate 3D representations, \textit{“2D MLLMs”} use only RGB inputs, without any explicit 3D supervision but trained for specific tasks and \textit{"Zero-shot VideoLMs"} methods that do not train using 2D/3D information. Our method falls into this category, and we further report two variants of \ours{}: w/o \textbf{DAS} and w/o \textbf{V-Rank}. On both benchmarks, \ours{} achieves the highest overall score with \textbf{94.8} CIDEr and \textbf{48.4} ROUGE scores on ScanQA and \textbf{57.2} EM on SQA3D, outperforming methods that consume explicit 3D information as well as video-based methods that require task-specific fine-tuning or extra auxiliary information. 
\begin{table}[t]
\centering
\scriptsize
\setlength{\tabcolsep}{2.8pt}
\caption{\textbf{Semantic judge and Embodied QA.} Left: semantic metrics on ScanQA (val) under the \judge{} judge. Right: transfer~to~OpenEQA.}
\label{tab:semjudge_main}\label{tab:eqa_transfer}
\begin{tabular}{@{}lcccc cc@{}}
\toprule
\multirow{2}{*}{\textbf{Method}} & \multicolumn{4}{c}{\textbf{ScanQA (val), semantic}} & \multicolumn{2}{c}{\textbf{OpenEQA}} \\
\cmidrule(lr){2-5}\cmidrule(l){6-7}
& \textbf{JC}$\uparrow$ & \textbf{JCP}$\uparrow$ & \textbf{SR}$\uparrow$ & \textbf{\judge{}}$\uparrow$ & \textbf{LLM-M.}$\uparrow$ & \textbf{Tok-F1}$\uparrow$ \\
\midrule
LLaVA-OV \textbf{(Baseline)} & 24.15 & 35.85 & 11.70 & 30.00 & 42.72 & 28.98 \\
\rowcolor{ourslight}
\ours{} w/o DAS    & 29.80 & 41.85 & 12.05 & 35.83 & 44.18 & 32.41 \\
\rowcolor{ourslight}
\ours{} w/o V-Rank & 28.65 & 39.95 & 11.30 & 34.30 & 44.47 & 31.96 \\
\rowcolor{oursmid}
\textbf{\ours{}} & \textbf{39.25} & \textbf{52.77} & \textbf{13.52} & \textbf{46.01} & \textbf{46.60} & \textbf{33.93} \\
\bottomrule
\end{tabular}
\end{table}

The ablated variants further show that both \textbf{DAS} and \textbf{V-Rank} contribute to the final gain. Detailed tool-wise analysis is in Supplementary Section~F, backbone comparisons in Supplementary Section~H and more results in Supplementary Section~C.\\
\textbf{Results with \texttt{SEU} Judge.} Table~\ref{tab:semjudge_main} verifies that the gains of \ours{} persist under a wider semantic context, which credits meaning-preserving answers needed for spatial understanding. Relative to the LLaVA-OV baseline, \ours{} improves JC from 24.15 to 39.25 \textbf{(+15.1)}\%, JCP from 35.85 to 52.77 \textbf{(+16.9)}\%, SR from 11.70 to 13.52 \textbf{(+1.82\%)}, and \judge{} from 30.0 to 46.01 \textbf{(+16.01\%)}. Overall, the judge widens the scope and validation for complex queries and demonstrates the ability of \ours{} in open-ended response generation.

\subsection{\ours{}: Dense Captioning and \mbox{Embodied QA}}
\label{sec:exp_transfer}
\begin{table}[t]
\centering
\scriptsize
\setlength{\tabcolsep}{3.6pt}
\renewcommand{\arraystretch}{1.05}
\caption{\textbf{Dense captioning transfer on Scan2Cap.}
We report results on Scan2Cap$_{\mathrm{val}}$ at IoU@0.5.}
\label{tab:densecap_transfer}
\begin{tabular}{@{}lcccc@{}}
\toprule
\multirow{2}{*}{\textbf{Method}} & \multicolumn{4}{c}{\textbf{Scan2Cap}$_{\mathrm{val}}$ \textbf{[IoU@0.5]}} \\
\cmidrule(l){2-5}
& \textbf{ROUGE}$\uparrow$ & \textbf{BLEU-4}$\uparrow$ & \textbf{METEOR}$\uparrow$ & \textbf{CIDEr}$\uparrow$ \\
\midrule

\rowcolor{taskblue}
\multicolumn{5}{@{}l@{}}{\textbf{\textit{Task-Specific}}} \\
Scan2Cap~\cite{chen2021scan2cap}      & 44.5 & 23.3 & 22.0 & 35.2 \\
3DJCG~\cite{cai20223djcg}             & 50.8 & 31.0 & 24.2 & 49.5 \\
3D-VLP~\cite{jin2023context}          & 51.5 & 32.3 & 24.8 & 54.9 \\
3D-VisTA~\cite{zhu20233d}             & 54.3 & 34.0 & 26.8 & 61.6 \\

\midrule
\rowcolor{llmpink}
\multicolumn{5}{@{}l@{}}{\textbf{\textit{3D LMMs}}} \\
LL3DA~\cite{chen2024ll3da}            & 55.1 & 36.8 & --   & --   \\
LEO~\cite{huang2023embodied}          & 58.1 & 38.2 & 27.9 & 72.4 \\
ChatScene~\cite{huang2023chat}        & 58.1 & 36.3 & 28.0 & 77.1   \\

\midrule
LLaVA-OV \textbf{(Baseline)}          & 52.2 & 33.5 & 25.2 & 58.9 \\

\ourcelllight{\ours{} w/o DAS} &
\ourcelllight{61.7} &
\ourcelllight{39.8} &
\ourcelllight{29.3} &
\ourcelllight{79.6} \\

\ourcellmid{\ours{} w/o V-Rank} &
\ourcellmid{62.0} &
\ourcellmid{40.0} &
\ourcellmid{29.5} &
\ourcellmid{80.1} \\

\ourcelldark{\textbf{\ours{}}} &
\ourcelldark{\textbf{62.3 {\scriptsize\textcolor{darkgreen}{$\uparrow$4.2}}}} &
\ourcelldark{\textbf{40.3 {\scriptsize\textcolor{darkgreen}{$\uparrow$2.1}}}} &
\ourcelldark{\textbf{29.8 {\scriptsize\textcolor{darkgreen}{$\uparrow$1.8}}}} &
\ourcelldark{\textbf{80.8 {\scriptsize\textcolor{darkgreen}{$\uparrow$3.7}}}} \\

\bottomrule
\end{tabular}
\end{table}
\textbf{Dense Captioning.} Next, we evaluate our pipeline for dense captioning without any additional tools or fine-tuning. As shown in Tab.~\ref{tab:densecap_transfer}, \ours{} beats the baseline and SOTA methods by an average of over \textbf{5}\% across all metrics CIDEr, BLEU, METEOR, and ROUGE-L. 
This indicates that selecting target-relevant RGB information also benefits the generation of object- and scene-level descriptions, often with accurate attributes.\\
\textbf{Embodied QA.} As shown in Tab.~\ref{tab:eqa_transfer}, \ours{} also transfers to Embodied QA and improves over the backbone model~\cite{li2024llava} by a significant margin. This highlights the strength of our pipeline and its ability to generalize to other~tasks~effectively. 
\begin{figure*} [!htb]
    \centering
    \includegraphics[width=0.95\linewidth]{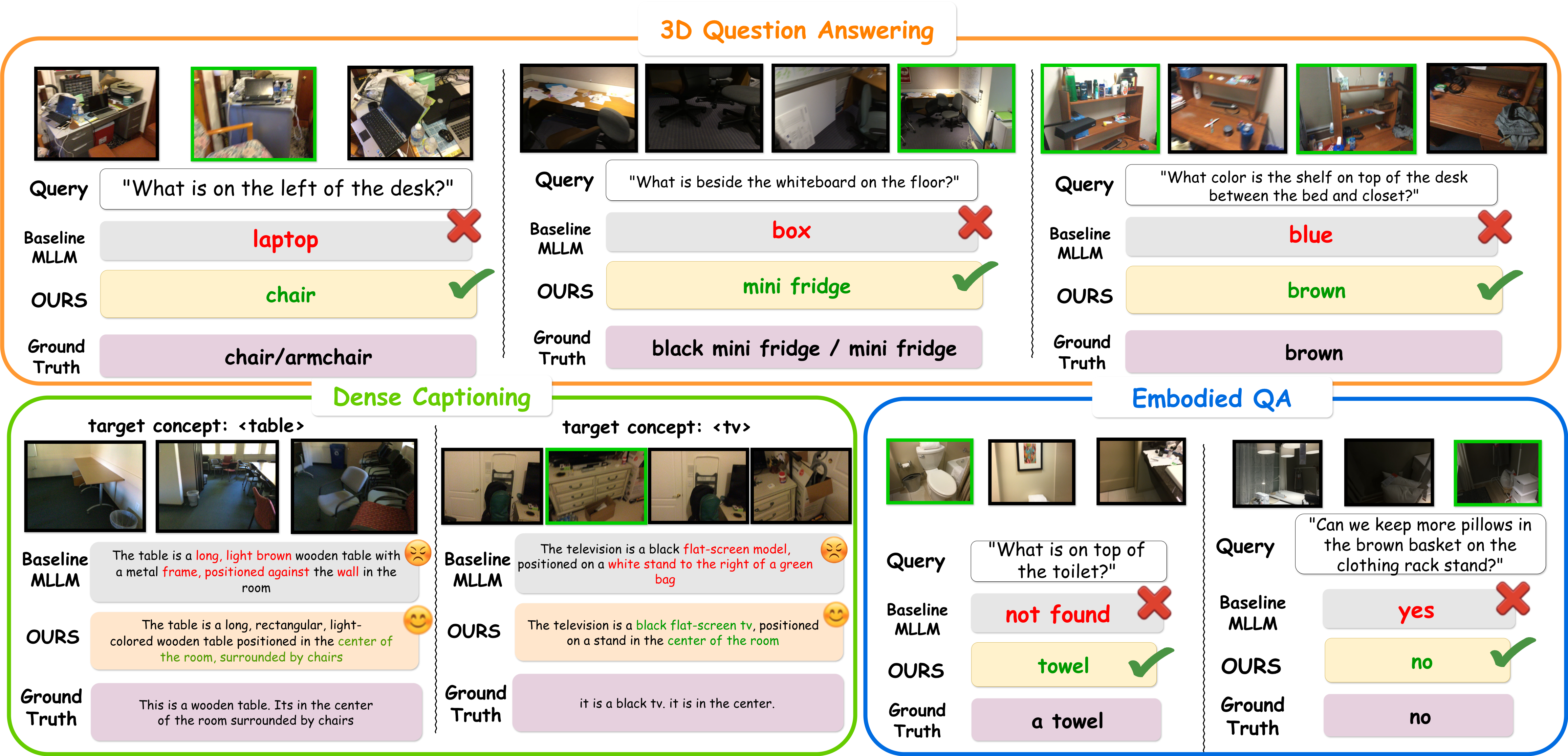}
    \caption{\textbf{Main Results of \ours{}.} \textbf{Top:} We showcase our results on 3D QA. \textbf{Bottom Left:} Performance on dense captioning \textbf{Bottom Right:} Performance of \ours{} on EQA.}
    \label{fig:main_result}
\end{figure*}

\section{Discussion}
\label{sec:exp_ablation}
In this section, we discuss how each component of \ours{} increases effectiveness. Here, we primarily look into different view selection strategies, scaling performance, and tools that help the VLM increase its \mbox{response capabilities}.

\begin{figure}[t]
\centering
\resizebox{0.75\linewidth}{\height}{\includegraphics[width=0.5\linewidth,trim={680 0 0 0},clip]{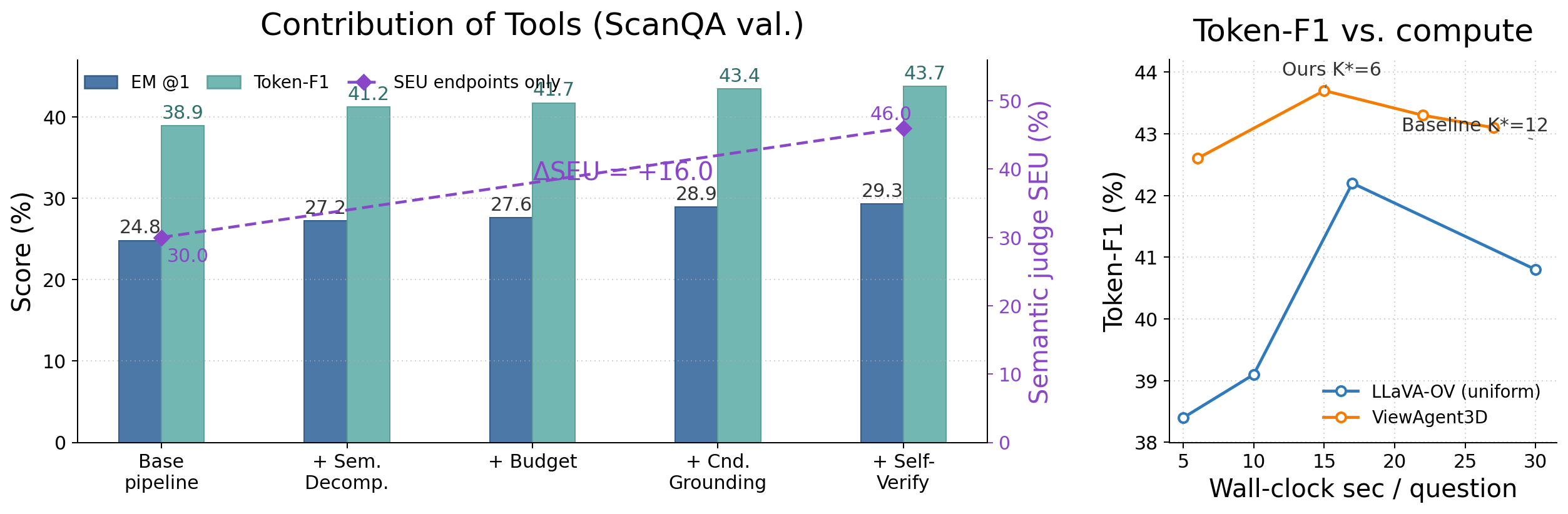}}
\caption{\textbf{View-budget scaling.} Token-F1 on ScanQA (val) against wall-clock time per question as the number of views grows.}
\label{fig:scaling}
\end{figure}
\textbf{Views and scaling.} Fig.~\ref{fig:scaling} shows the effect of increasing the number of views on the inference time. \ours{} reaches the optimal point of performance at less than \textbf{50}\% of the baseline's budget, while the baseline with more views still performs worse, underscoring the importance of proper evidence selection and reasoning within the fixed context of the model.

\begin{table}[t]
\centering
\footnotesize
\setlength{\tabcolsep}{6pt}
\renewcommand{\arraystretch}{0.85}
\caption{\textbf{Tool Usage Analysis.}
In each row we examine the cumulative effect of progressively enabling each test-time tool on the ScanQA (val) dataset.}
\label{tab:evidence_control}
\begin{tabular}{@{}lcc@{}}
\toprule
\textbf{Tool} & \textbf{EM}$\uparrow$ & \textbf{Token-F1}$\uparrow$ \\
\midrule
Base VLM + view selection
& 24.8
& 39.2 \\
+ Sem. Concepts.
& 27.2 {\scriptsize\textcolor{darkgreen}{$\uparrow$2.4}}
& 41.2 {\scriptsize\textcolor{darkgreen}{$\uparrow$2.0}} \\
+ Budget
& 27.6 {\scriptsize\textcolor{darkgreen}{$\uparrow$2.8}}
& 41.5 {\scriptsize\textcolor{darkgreen}{$\uparrow$2.3}} \\
+ Cnd. Grounding
& 28.9 {\scriptsize\textcolor{darkgreen}{$\uparrow$4.1}}
& 43.1 {\scriptsize\textcolor{darkgreen}{$\uparrow$3.9}} \\
+ Self-Verify
& 29.3 {\scriptsize\textcolor{darkgreen}{$\uparrow$4.5}}
& 43.7 {\scriptsize\textcolor{darkgreen}{$\uparrow$4.5}} \\
\midrule
\rowcolor{oursmid}
\textbf{\ours{}}
& \textbf{29.3}
& \textbf{43.7} \\
\bottomrule
\end{tabular}
\end{table}
\textbf{Effect of Tools.} Tab.~\ref{tab:evidence_control} measures the effect of each tool on the overall pipeline. Keeping the base pipeline (with proper view selection), we see that concept extraction (\textbf{2.4}\%) and grounding (\textbf{4.1}\%) help significantly in improving performance as they help in increasing the quality of retrieved views and localization of proper evidence. Proper views and verification further helps constrain the \mbox{answer's} search space and generate refined responses.(more in \mbox{Supplementary Section~F})\\
\begin{table}[t]
\centering
\footnotesize
\caption{\textbf{View selection strategies.}
Compared to the baseline, we progressively add random or uniform coverage, semantic retrieval (\emph{SemRet}), DAS, and V-Rank. Green and red margins denote respective changes in performance over Row~\circnum{1}.}
\label{tab:view_selection}
\setlength{\tabcolsep}{1.7pt}
\renewcommand{\arraystretch}{0.90}
\begin{tabular}{@{}clcccc@{}}
\toprule
& \multirow{2}{*}{\textbf{Method}} 
& \multicolumn{3}{c}{\textbf{ScanQA}} 
& \multicolumn{1}{c}{\textbf{SQA3D}} \\
\cmidrule(lr){3-5}\cmidrule(l){6-6}
\textbf{\#} & 
& \textbf{CIDEr}$\uparrow$ 
& \textbf{ROUGE-L}$\uparrow$ 
& \textbf{EM}$\uparrow$ 
& \textbf{EM}$\uparrow$ \\
\midrule

\circnum{1} & LLaVA-OV \textbf{(Baseline)}
& 50.0 & 29.4 & 15.6 & 25.4 \\

\circnum{2} & \circnum{1} + rand. views
& 46.8 {\scriptsize\textcolor{red!75!black}{$\downarrow$3.2}}
& 26.7 {\scriptsize\textcolor{red!75!black}{$\downarrow$2.7}}
& 10.8 {\scriptsize\textcolor{red!75!black}{$\downarrow$4.8}}
& 22.6 {\scriptsize\textcolor{red!75!black}{$\downarrow$2.8}} \\

\circnum{3} & \circnum{1} + uniform views
& 52.4 {\scriptsize\textcolor{darkgreen}{$\uparrow$2.4}}
& 30.8 {\scriptsize\textcolor{darkgreen}{$\uparrow$1.4}}
& 16.4 {\scriptsize\textcolor{darkgreen}{$\uparrow$0.8}}
& 27.9 {\scriptsize\textcolor{darkgreen}{$\uparrow$2.5}} \\

\circnum{4} & \circnum{3} + SemRet
& 79.1 {\scriptsize\textcolor{darkgreen}{$\uparrow$29.1}}
& 42.3 {\scriptsize\textcolor{darkgreen}{$\uparrow$12.9}}
& 23.8 {\scriptsize\textcolor{darkgreen}{$\uparrow$8.2}}
& 46.5 {\scriptsize\textcolor{darkgreen}{$\uparrow$21.1}} \\

\circnum{5} & \circnum{4} + DAS
& 88.6 {\scriptsize\textcolor{darkgreen}{$\uparrow$38.6}}
& 45.9 {\scriptsize\textcolor{darkgreen}{$\uparrow$16.5}}
& 26.9 {\scriptsize\textcolor{darkgreen}{$\uparrow$11.3}}
& 52.8 {\scriptsize\textcolor{darkgreen}{$\uparrow$27.4}} \\

\circnum{6} & \circnum{5} + V-Rank
& \textbf{94.8} {\scriptsize\textcolor{darkgreen}{$\uparrow$44.8}}
& \textbf{48.4} {\scriptsize\textcolor{darkgreen}{$\uparrow$19.0}}
& \textbf{29.3} {\scriptsize\textcolor{darkgreen}{$\uparrow$13.7}}
& \textbf{57.2} {\scriptsize\textcolor{darkgreen}{$\uparrow$31.8}} \\

\rowcolor{oursmid}
& \textbf{\ours{}}
& \textbf{94.8}
& \textbf{48.4}
& \textbf{29.3}
& \textbf{57.2} \\
\bottomrule
\end{tabular}
\end{table}%
\textbf{View selection strategies.} Tab.~\ref{tab:view_selection} compares different strategies for selecting views. Selecting diverse, retrieved set of views and reasoning over them (SemRet, DAS, and V-Rank) leads to much stronger gains (\textasciitilde\textbf{20}\%) than simply adding random or uniform views. 




\section{Conclusion}
\label{sec:conclusion}
In this work, we presented \ours{}, a training-free, test-time reasoning pipeline for 3D scene understanding over multi-view RGB images. Instead of constructing an explicit 3D representation, \ours{} treats 3D reasoning as a problem of selecting and reasoning over the right evidence: it decomposes the query, retrieves and diversifies candidate views, conditionally invokes 2D grounding, and answers from the selected observations. Across 3D QA, dense captioning, and embodied QA, this simple design remains competitive with more representation-heavy 3D pipelines, while our analyzes show that the main gains come from selecting better views rather than simply using more of them. Overall, our results suggest that 3D scene understanding can be performed from 2D observations alone when frozen models are given the right information.

\label{viewagent:references-start}
{\small
  \bibliographystyle{ieeenat_fullname}
  \bibliography{main}
}
\end{document}